%% file: main.tex
\documentclass{article}
\usepackage{spconf,amsmath,graphicx}
\usepackage{times}
\usepackage{epsfig}
\usepackage{amssymb}
\usepackage{xcolor}
\usepackage{url}
\usepackage[linesnumbered,ruled,vlined]{algorithm2e}
\usepackage[noend]{algpseudocode}
\usepackage{comment}
\usepackage{caption}
\usepackage{multirow}
\definecolor{vyellow}{rgb}{0.7490,0.5647,0}
\definecolor{vyelloworig}{rgb}{1,0.7529,0}
\definecolor{vmagenta}{rgb}{0.4392,0.1882,0.6274}
\definecolor{vmagenta3}{rgb}{0.7254,0.6352,0.7960}
\definecolor{vmagentaorig}{rgb}{0.9215,0.8039,1}
\definecolor{vblue}{rgb}{0.3686,0.7921,0.8431}
\definecolor{vred}{rgb}{0.7529,0,0}

\renewcommand{\paragraph}[1]{\vspace{1mm} \noindent \textbf{#1}}

\title{ARBITRARY POINT CLOUD UPSAMPLING VIA DUAL BACK-PROJECTION NETWORK}
\name{Zhi-Song Liu\thanks{corresponding email: zhisong.liu@dell.com}, Zijia Wang, Zhen Jia}
\address{Dell Research}
\begin{document}
%
\maketitle
\begin{abstract}
Point clouds acquired from 3D sensors are usually sparse and noisy. Point cloud upsampling is an approach to increase the density of the point cloud so that detailed geometric information can be restored. In this paper, we propose a Dual Back-Projection network for point cloud upsampling (DBPnet). A Dual Back-Projection is formulated in an up-down-up manner for point cloud upsampling. It not only back projects feature residues but also coordinates residues so that the network better captures the point correlations in the feature and space domains, achieving lower reconstruction errors on both uniform and non-uniform sparse point clouds. Our proposed method is also generalizable for arbitrary upsampling tasks (e.g. 4$\times$, 5.5$\times$). 
Experimental results show that the proposed method achieves the lowest point set matching losses with respect to the benchmark. In addition, the success of our approach demonstrates that generative networks are not necessarily needed for non-uniform point clouds.
\end{abstract}
%
\section{Introduction}

A point cloud is one of the popular ways to represent 3D surfaces because they capture high-frequency geometric information without requiring much memory. Recently, the study on point-cloud-based 3D analysis is getting more attention. A fundamental problem is, however, that the 3D sensors used in robotics or autonomous cars can only produce sparse, incomplete, or noisy point cloud data. For small objects or objects far from the camera, the collected point cloud cannot be directly used for analysis. The objective of this work is, given a sparse and noisy point cloud, to upsample it as dense ones by learning or non-learning approaches.
\begin{figure}[t]
	\vskip 0.01in
	\begin{center}
		\centerline{\includegraphics[width=0.95\columnwidth]{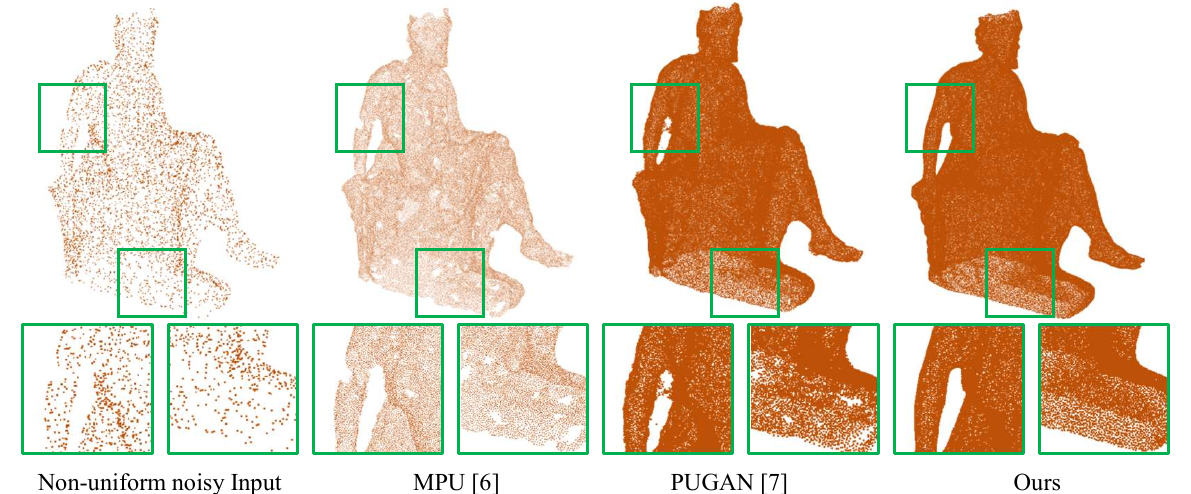}}
		\caption{16$\times$ upsampling from noisy non-uniform points.}
		\label{Figure 1}
	\end{center}
	\vskip -0.5in
\end{figure}
Recently, learning based point cloud upsampling methods~\cite{punet,mpu,pugan,pugeo} show that learning priors from data is a very promising direction. Although these deep-learning-based methods already achieved good progress on this task, there are still many issues to be solved: 1) developing a universal up-sampling model applicable to arbitrary upsampling tasks would be highly desirable, and 2) point cloud from real life is usually non-uniformly distributed and noisy. An ideal point cloud upsampling model should be robust against these different types and levels of artifacts.
 
 To address these two challenges, we introduce a novel, Dual Back-Projection network (DBPnet), specially designed for point cloud upsampling. Its originality is to tackle both coordinate and feature refinement using back projection. As shown in Figure~\ref{Figure 1}, in contrast with existing methods~\cite{mpu,pugan}, our new method is able to upsample noisy non-uniform distributed point clouds into dense, uniformly distributed ones. More importantly, we train our proposed network for a large upsampling factor, e.g., 16$\times$, and downsample it to achieve arbitrary upsampling. We claim the following original contributions 1) we introduce a novel, dual back-projection process to learn point correlations for upsampling tasks. Information from both feature and coordinate spaces are updated using an up-down-up process. 2) Position-aware attention is embedded into the back-projection process to learn non-local point correlations. With the help of this positional embedding, the attention module can incorporate both position and feature correlations to capture global representation. 3) A resampling stage is introduced as a second-stage point cloud refinement where dense points are downsampled to sparse point sets. Hence, each generated point gets a chance to attend to information from the neighborhoods of the original sparse point cloud. This stage acts as a self-supervision where local and global information are combined for coordinate estimation.

\section{Related Work}
\noindent\textbf{Point cloud Upsampling} Early work in the Computer Graphics field used optimization methods to solve point cloud upsampling problems~\cite{opt_1,opt_2,opt_3,opt_4}. For example,~\cite{opt_1} assumed that the new inserted points should lay on a smooth surface so that they can be optimized by the moving least squares. To ensure sharp reconstruction, Huang et al.~\cite{opt_4} proposed a progressive method called EAR for edge-aware resampling of points. Deep learning techniques recently spread in a variety of fields, including 3D processing. The first work on point cloud upsampling was PUNet~\cite{punet}. The authors proposed a convolutional neural network to directly extract features from point sets. To upsample the points, the learned features were expanded and separately mapped to a subset of the point cloud. To avoid point clustering, repulsion loss was used to balance the point distances. To explicitly upsample the point cloud, Wang et al.~\cite{mpu} proposed to use 1\textit{d} code to attach features to different locations. By stacking multiple 2$\times$ upsampling blocks, their method can progressively upsample the point cloud to the desired number of points. As mentioned in~\cite{punet}, upsampling points is equivalent to upsampling features. Li et al.~\cite{pugan} resolve it by using a generative adversarial network to learn point distribution from the latent space. Most recently, SSPU~\cite{sspu} utilizes multi-view rendering to supervise the dense point cloud generation. SAPCU~\cite{sapcu} seeks to find rich points from a surface learned from neural implicit functions. PU-GCN~\cite{pu-gcn} designs a graphic convolution network to better encode coordinate information for upsampling.

\begin{figure}[t]
	\vskip 0.01in
	\begin{center}
		\centerline{\includegraphics[width=\columnwidth]{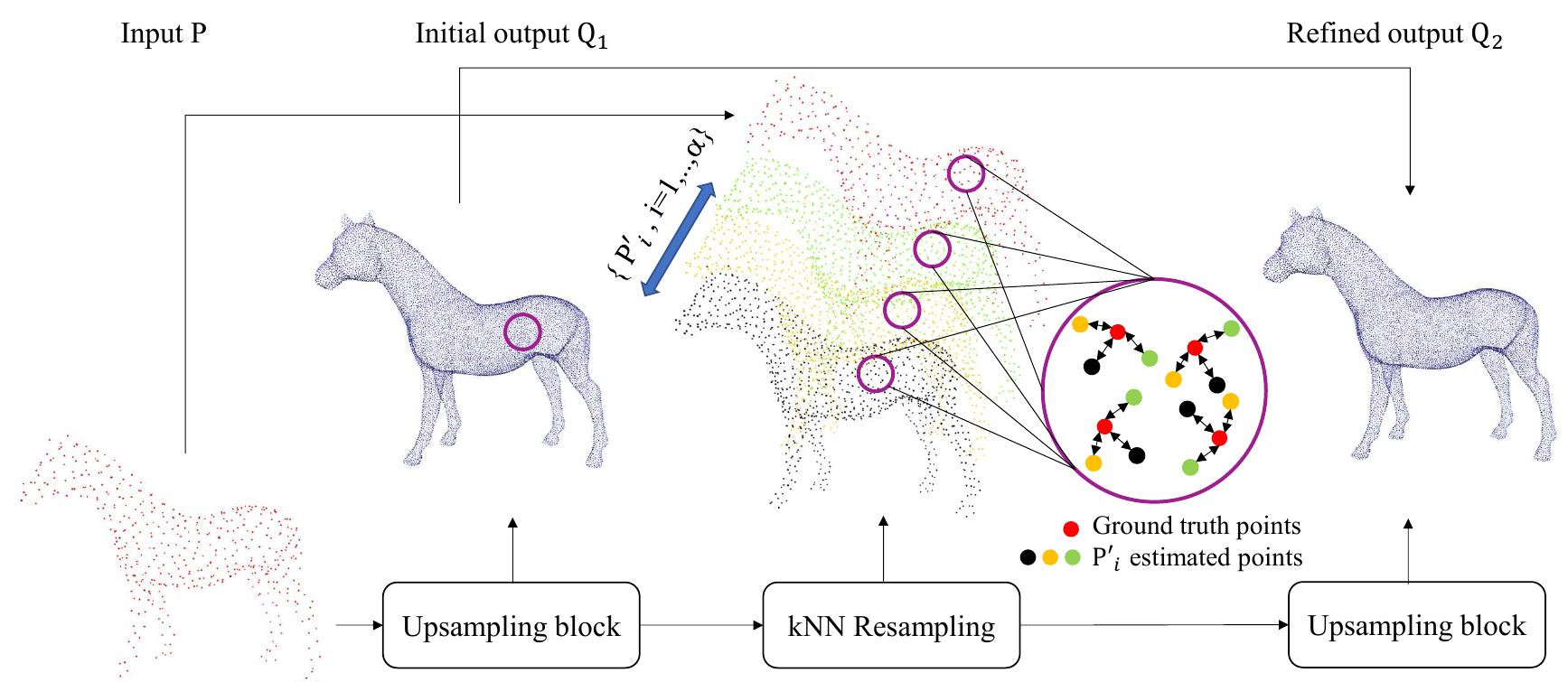}}
		\caption{Overview of our proposed DBPnet model. It includes two stages of $\alpha\times$ upsampling processes. The upsampling block contains feature-based back-projection for feature expansion. The initial upsampled points are resampled by kNN to obtain $\alpha$ subsets of estimated points. Along with the input points, the second stage of upsampling process works as a coordinate-based back-projection to update the residual distance between input points and neighborhood points.}
		\label{Figure 2}
	\end{center}
	\vskip -0.5in
\end{figure}

\section{Method}
\begin{figure}[t]
	\vskip 0.01in
	\begin{center}
		\centerline{\includegraphics[width=\columnwidth]{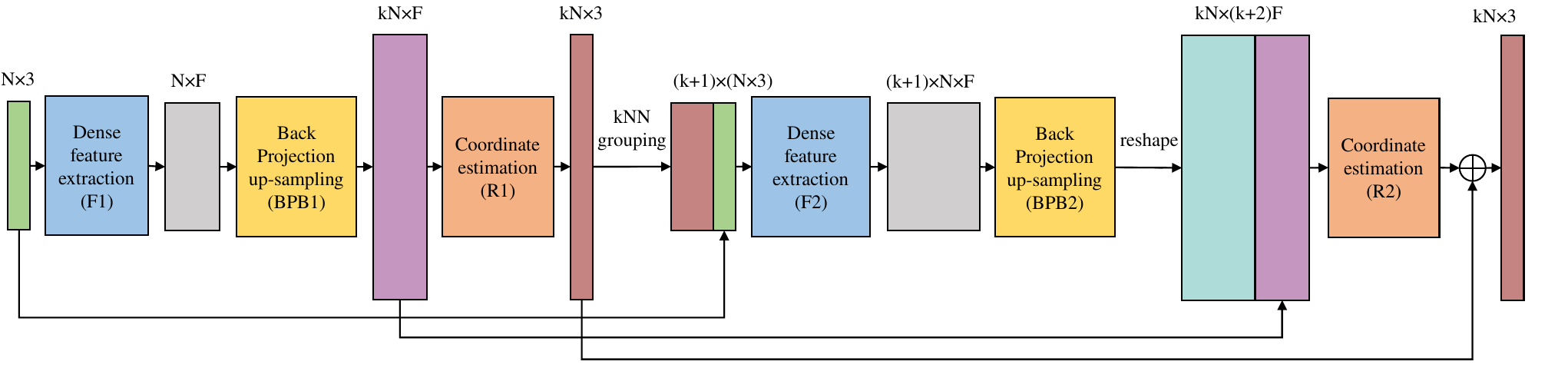}}
		\caption{The proposed DBPnet model, including Deep feature extraction, Back-Projection up-sampling and Coordinate estimation.}
		\label{Figure 3}
	\end{center}
	\vskip -0.4in
\end{figure}

Given a sparse non-uniform 3D point cloud $P={p_i}_{i=1}^N$, our network aims to generate dense point cloud $Q={q_i}_{i=1}^{\alpha N}$, where $\alpha$ is the upsampling factor. The generated point cloud should follow the same underlying surface as the target object and be uniformly distributed. As shown in Figure~\ref{Figure 2} and Figure~\ref{Figure 3}, our proposed Dual Back-Projection network (DBPnet) consists of three components: 1) Dense feature extraction, 2) Back-Projection upsampling, and 3) Coordinate estimation to achieve a two-step upsampling process. The overall process is a coordinate-based back-projection. The initial upsampled point sets are downsampled again to match with the input point sets, and the residual distances computed using the k nearest neighbors are back-projected for refinement. The refined and the initial point sets are combined together for the final coordinate estimation. Within the upsampling block, the feature-based back-projection iteratively updates point features for estimation. Combining coordinate and feature-based back-projection, we achieve significant improvements over previous works. 

\noindent\textbf{Motivation} The overall upsampling network is built upon the coordinate-based back-projection, which can be described using the following equation:
\begin{small}
	\begin{equation}
	Q^*=f\bigg(\mathcal{S}\{g[f(P)] \setminus P\}\bigg)\cap f(P) \tag{1}
	\label{Equation 1}
	\end{equation}
\end{small}
\noindent 
where \textit{f}($\cdot$) is the upsampling operation, \textit{g}($\cdot$) the resampling operation, $\setminus$ is the set-difference operation, $\mathcal{S}$ is the dimension switch operation, $P\in\mathbb{R}^{N\times3}$  is the input sparse points and $Q^*\in\mathbb{R}^{\alpha N\times3}$ is the upsampled points. We omitted feature extraction and coordinate reconstruction for simplicity.

Following the process of back projection, we first upsample the input points and then resample them to the same dimension as the input. The resampling process is different from the downsampling process in the sense that we do not sample only one sparse point subset but rather gather the points into a set of subgroups. For example, we can find \textit{N} points as cluster centroids and then use \textit{k} Nearest Neighborhood (kNN) to group neighboring points into these clusters. As shown in Figure~\ref{Figure 2}, the upsampled points \textit{P} can be resampled into $\alpha$ sparse point subsets. Together with the input point set, we form a pool of sparse point sets that can be used for neighborhood feature extraction. Hence we learn the relative complement of the set \textit{P} with respect to the set \textit{Q}. As mentioned in PUNet~\cite{punet}, features and points are interchangeable, so feature expansion is equivalent to expanding the number of points. After resampling, we thus expand features and then switch the dimensions of features and points. For $\alpha\times$ upsampling, we use $\alpha+1$ features (one point from the input set plus $\alpha$ resampled points for reconstruction. There are three merits of using coordinate-based back-projection: 1) it forms a global loop of self-supervision where the target points should locate around the input points, achieving kNN interpolation, 2) it avoids the expensive computation of kNN searches and feature extraction in every layer, and 3) by using resampling, it ensures each point collects both local and global information for estimation.

\begin{figure}[t]
	\vskip 0.01in
	\begin{center}
		\centerline{\includegraphics[width=\columnwidth]{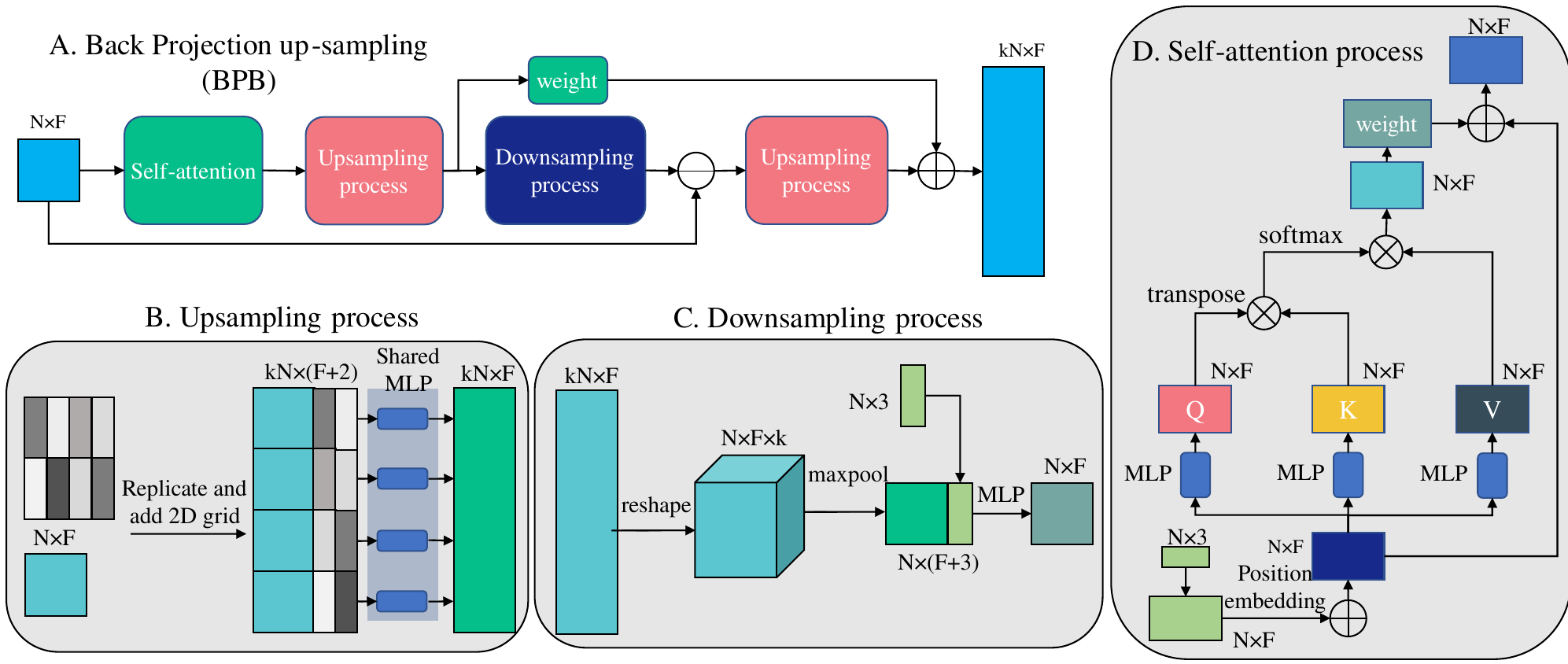}}
		\caption{Back-Projection upsampling. It includes up-sampling, down-sampling and self-attention blocks. It iteratively up-down-up upsamples the point features.}
		\label{Figure 5}
	\end{center}
	\vskip -0.5in
\end{figure}

\noindent\textbf{Back-Projection Upsampling.} The key component of the upsampling block is the use of back-projection. We call it feature-based back-projection because it is done in the feature domain. As shown in Figure~\ref{Figure 5}, it includes three components, namely upsampling, downsampling, and self-attention. The basic procedure is shown in Figure~\ref{Figure 5}A. The input sparse point features are firstly upsampled to the dense features, and then the dense features are downsampled and subtracted from the input features to calculate residues. The residual information is further upsampled for updating the dense point features. 

\noindent\textbf{Position-aware self-attention.} We strive for a better feature representation for point cloud upsampling. To achieve this goal, attention is a good choice because of its nonlocal learning ability. It has been widely used in many fields~\cite{atten_1,atten_2,atten_3,atten_4}, including 3D point cloud~\cite{atten_6,atten_7,atten_8,atten_9}. What distinguishes our position-aware self-attention from other works is that we introduce the position embedding technique for attention computation. As introduced in~\cite{attention}, adding position codes in the attention computation can incorporate the order information into the model. Though the point cloud is unordered data that is invariant to permutation, the relative position defines the structure of the underlying 3D surface. It is necessary to keep the position information, that is, coordinate data for feature extraction. To this end, we add \textit{positional encoding} to the point feature to retain positional information. It can also be understood as a Laplacian theory \textit{L=D-A}, where \textit{L} is the Laplacian matrix, \textit{D} is the degree matrix and \textit{A} is the adjacency matrix. Position-aware attention is equivalent to replacing \textit{D} by \textit{positional encoding} and replacing \textit{A} by point features. Their combination forms the matrix representation of a graph. By using deep learning, both degree and adjacency matrices can be learned end-to-end. Mathematically, we can describe the process as follows,
\begin{small}
	\begin{equation} 
	\mathbf{y}=\omega\{softmax\left(\theta(\mathbf{z}\right)\phi(\mathbf{z}^T))\mathit{g}(\mathbf{z})\}+x \tag{2}
	\label{Equation 2}
	\end{equation}
\end{small}
where $z=x+E_{pos}$ is the position-aware encoding. It is the sum of point feature \textit{x} and \textit{positional encoding} $E_{pos}$. For attention computation, the input data are separately transformed by MLPs and multiplied for autocorrelation. We also add a skip-connection to further improve the learning ability, where $\omega$ is the weighting process to finetune the amount of attention values. 

\noindent\textbf{Loss functions}
To keep data fidelity, the full loss function is defined as $\mathcal{L}=\mathcal{L}_{CD} + \lambda \mathcal{L}_{uni}$, where $\lambda$ is the weighting factor used to balance the Chamfer loss ($\mathcal{L}_{CD}$)~\cite{mpu,pugan} with respect to the uniform loss ($\mathcal{L}_{uni}$)~\cite{punet}.

\begin{figure}[t]
	\vskip 0.01in
	\begin{center}
		\centerline{\includegraphics[width=\columnwidth]{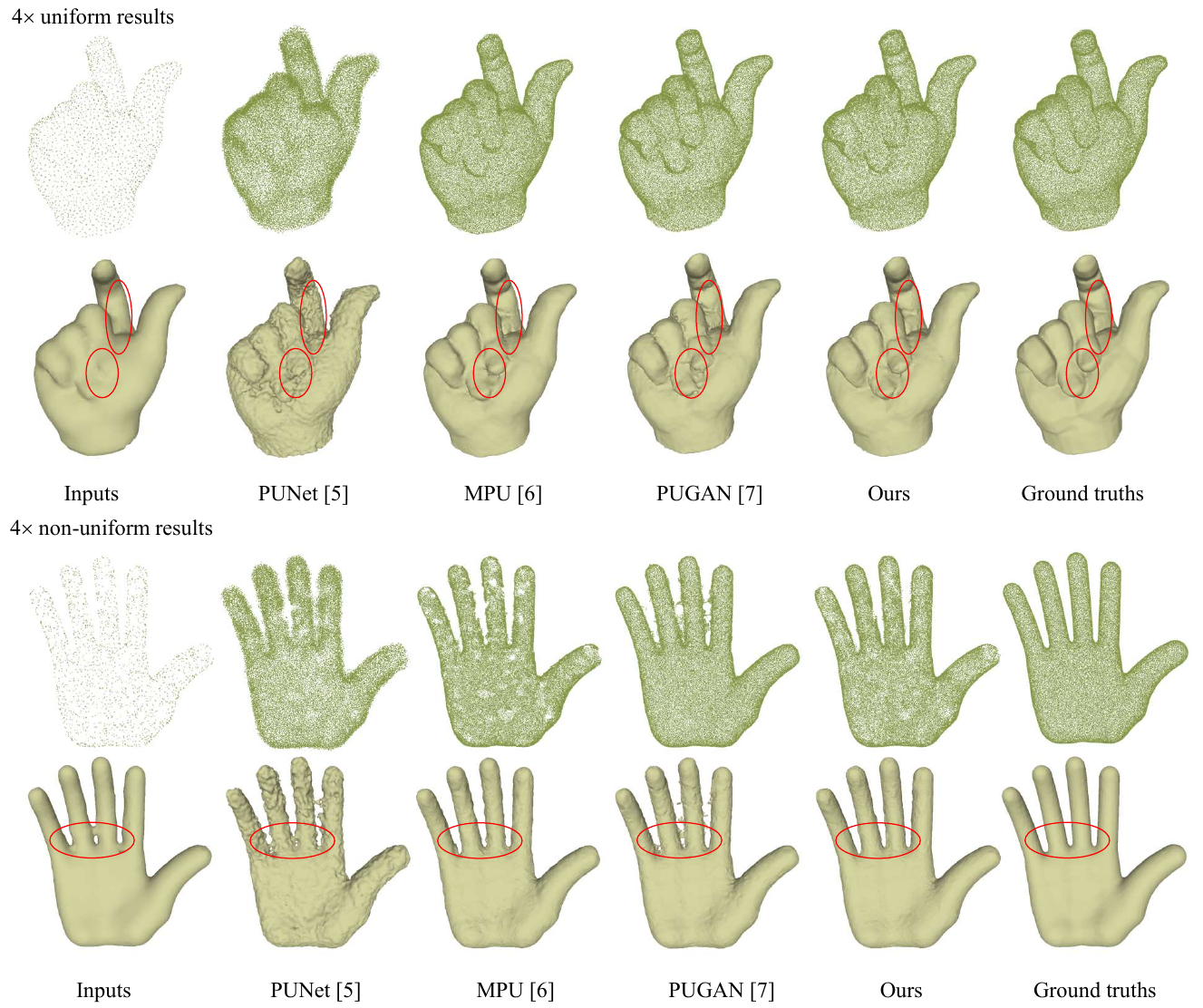}}
		\caption{4$\times$ upsampling results from 2048 input points (uniformly and non-uniformly) and reconstructed mesh.}
		\label{Figure 6}
	\end{center}
	\vskip -0.5in
\end{figure}

\section{Experiments}
\noindent\textbf{Training dataset and Implementation details.} For training data, we used the same database as PUNet~\cite{punet}, consisting of 60 different 3D models from the Visionary repository~\cite{data}. For each 3D model, we used farthest point sampling to select seeds. Then we employed the Poisson disk sampling~\cite{poisson} to generate target dense patches \textit{Q}. Different from ~\cite{pugeo,mpu}, we used random sampling to generate input sparse patches \textit{P} from \textit{Q}. The number of input and target patches is \textit{N} and $\alpha N$ with \textit{N}=256 and $\alpha$=16. Therefore, our training task is to map the non-uniformly distributed sparse \textit{P} to the uniformly distributed dense points \textit{Q}. Note that we only trained 16$\times$ upsampling model. The reason is that if a 16$\times$ result can represent the underlying surface well, a simple uniform downsampling process can faithfully represent the same 3D structure. Hence, it is not necessary to train the same model for smaller upsampling scenarios. We thus used the furthest sampling approach to downsample the 16$\times$ results, in order to achieve 2$\times$, 4$\times$ and 8$\times$ results, respectively. For testing data, we used the same data as PUGAN~\cite{pugan}, consisting of 27 different 3D models with a wide variety of simple and complex objects. From each model, 2048 random points were extracted to form the input sparse point cloud. To better demonstrate the robustness of our method, we tested four different scenarios: clean uniformly sampled points, clean non-uniformly sampled points, noisy non-uniformly sampled points and real points collected from LIDAR. Our comparison includes EAR~\cite{opt_4}, PUNet~\cite{punet}, MPU~\cite{mpu} and PUGAN~\cite{pugan}.
For evaluation, we use Chamfer distance (CD), Hausdorff Distance (HD), Point-to-Surface distance (P2F) mean value and uniformity value.

\input{tab/sota}

\noindent\textbf{Comparison with State-of-the-art Methods.} Table~\ref{tab:sota} summarizes the quantitative comparison results. We have two sets of input data: uniform (input points are evenly distributed) and non-uniform (input points are randomly distributed). Our DBPnet achieves the best performance with the lowest values consistently for different evaluation metrics. Specifically, it can be observed that our new method achieves the lowest P2F and Uniformity values on every task, indicating that it achieves the generation of uniform, dense point cloud that remain close to the underlying surface. For visual comparison (Figure~\ref{Figure 6}), we applied surface reconstruction to the upsampled point sets by using PCA normal estimation (number of neighborhood=20)~\cite{pca} and screened Poisson reconstruction (depth=9)~\cite{screen}. Results show that our method is able to fill holes and robustly generate uniformly-sampled point sets, while former methods, such as EAR, PUNet and MPU, adequately handle uniform point sets but fail on non-uniformly sampled regions. 

\input{tab/ablation}

\noindent\textbf{Ablation studies.} As introduced in Section 2, the coordinate-based back-projection works as a second-stage refinement. To make a comparison, we define the \textit{baseline} as our network without the coordinate-based back-projection and feature-based back-projection blocks. Next, we define \textit{Feature BP} as the baseline network using feature-based back-projection for upsampling; \textit{Coordinate BP} as the baseline network using coordinate-based back-projection for upsampling; \textit{Position Embedding} as the baseline network using position-aware attention and \textit{full pipeline} as the complete proposed network. Table~\ref{tab:ablation} shows the evaluation results. The full pipeline performs the best. Removing any key components can reduce the performance, hence it indicates the contributions of each component. 

\begin{figure}[t]
	\vskip 0.01in
	\begin{center}
		\centerline{\includegraphics[width=0.9\columnwidth]{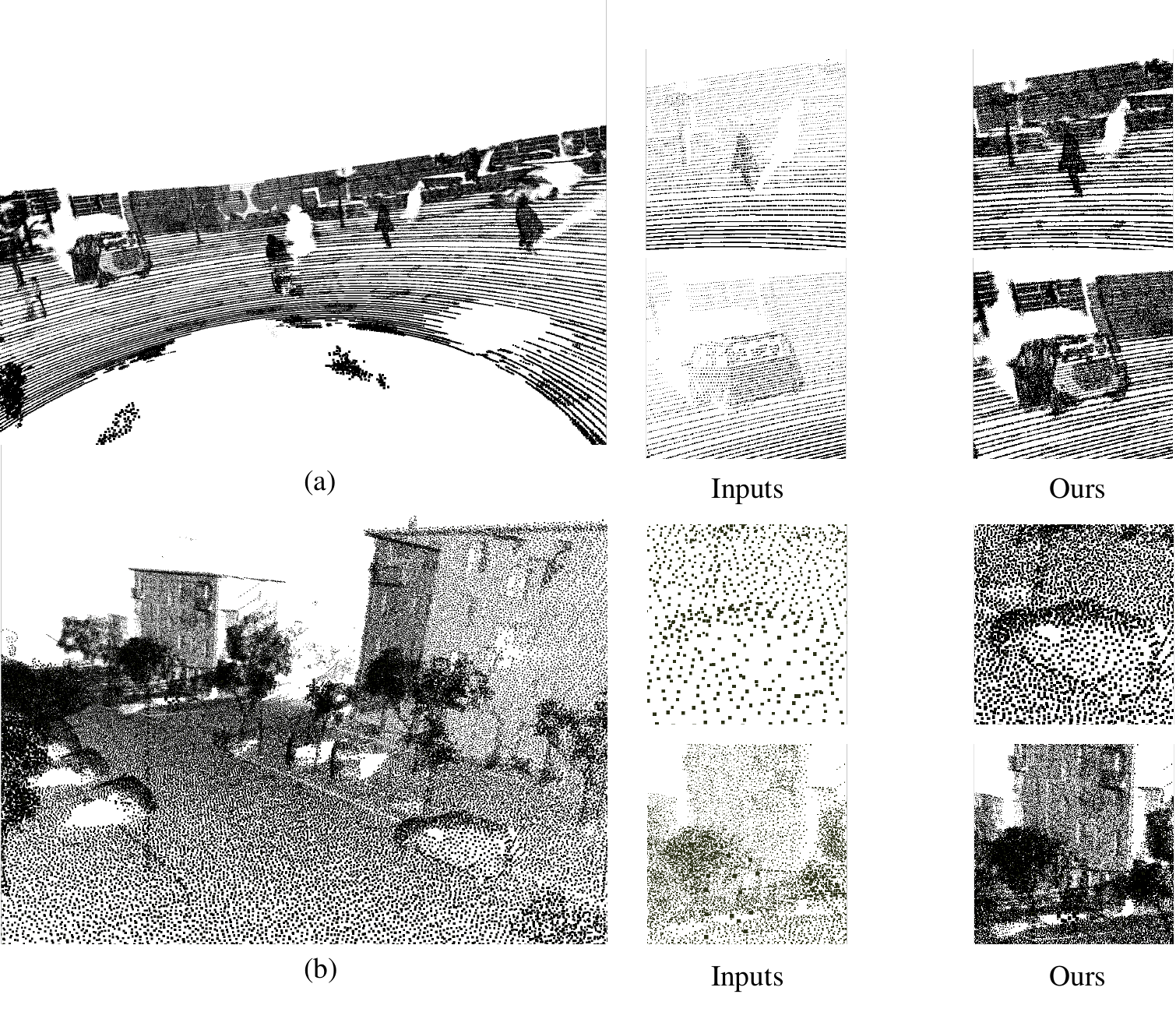}}
		\caption{16$\times$ upsampling results on real 3D points. (a) KITTI data~\cite{kitti} of the driving screen and (b) Paris-Lille-3D data~\cite{paris} of the city view of Paris.}
		\label{Figure 9}
	\end{center}
	\vskip -0.5in
\end{figure}

\noindent \textbf{Real World Point cloud.} To evaluate our model on real scans, we tested it on the KITTI~\cite{kitti} dataset and the Paris-Lille-3D~\cite{paris} dataset. Figure~\ref{Figure 9} It can be found that using our method can upsample the point sets to better represent objects, like cars and pedestrians.

\section{Conclusion}
In this work, we introduced a Dual Back-Projection network (DBPnet), specially designed for point cloud upsampling. Combining both Coordinate and Feature-based back-projections, this model is able to
reveal detailed geometric structures from sparse and noisy input point clouds. We also proposed a position-aware attention mechanism, 
to complement the non-local representation with positional information. Overall, we formed a network enabling both global coordinate refinement and local feature refinement. The such an adaptive patch-based network learns point distributions and is able to handle both uniformly and non-uniformly distributed point sets. Extensive experiments and studies demonstrated that the proposed solution outperforms state-of-the-art methods in both quantitative and qualitative comparisons.

{\small
\bibliographystyle{IEEEbib}
\bibliography{egbib}
}

\end{document}

%% file: tab/sota.tex
\begin{table}[t]
	\caption{Quantitative evaluation of state-of-the-art point clouds upsampling approaches for scales 2$\times$, 4$\times$, 8$\times$ and 16$\times$. {\color{red}Red} indicates the best.}
	\label{tab:sota}
	\vskip -0.05in
	\normalsize
	\begin{center}
                \resizebox{\linewidth}{!}{
			\begin{tabular}{cccccccccc}
				\hline
				\multicolumn{2}{c}{No. of sparse point clouds}                                       & \multicolumn{8}{c}{2048}                                                                                       \\ \hline
				\multicolumn{2}{c}{Sampling}                                           & \multicolumn{4}{c|}{Uniform}                                     & \multicolumn{4}{c}{Non-uniform}             \\ \hline
				\multicolumn{2}{c|}{Eval. ($10^{-3}$)}                                     & CD    & HD    & P2F & \multicolumn{1}{c|}{Uniformity} & CD    & HD    & P2F & Uniformity \\ \hline
				\multicolumn{1}{c|}{\multirow{5}{*}{2x}}  & \multicolumn{1}{c|}{EAR~\cite{opt_4}}   & 0.471 & 2.912 & 2.896     & \multicolumn{1}{c|}{1.749}           & 0.705 & 7.027 & 2.481     & 3.748           \\
				\multicolumn{1}{c|}{}                     & \multicolumn{1}{c|}{PUNet~\cite{punet}} & 0.822 & 4.089 & 4.974     & \multicolumn{1}{c|}{1.579}           & 1.009 & 7.331 & 9.776     & 2.296           \\
				\multicolumn{1}{c|}{}                     & \multicolumn{1}{c|}{PUGAN~\cite{pugan}} & 0.433     & 4.133     & 1.550         & \multicolumn{1}{c|}{1.189}               & 0.499     & 4.420     & 3.801         & 2.001               \\
				\multicolumn{1}{c|}{}                     & \multicolumn{1}{c|}{MPU~\cite{mpu}}   & 0.446 & 2.096 & 1.546     & \multicolumn{1}{c|}{1.937}           & 0.712 & 6.234 & 1.968     & 3.234           \\
				\multicolumn{1}{c|}{}                     &
				\multicolumn{1}{c|}{PU-GCN~\cite{pu-gcn}}   & {\color{red}0.355} & {\color{red}1.967} & 1.533     & \multicolumn{1}{c|}{1.870}           & 0.660 & 5.742 & {\color{red}1.955}     & 3.077           \\
				\multicolumn{1}{c|}{}                     &
				\multicolumn{1}{c|}{Ours}  & 0.409 & 2.904 & {\color{red}1.531}     & \multicolumn{1}{c|}{{\color{red}1.154}}           & {\color{red}0.426} & {\color{red}3.762} & {\color{red}1.966}     & {\color{red}1.166}           \\ \hline
				\multicolumn{1}{c|}{\multirow{5}{*}{4x}}  & \multicolumn{1}{c|}{EAR~\cite{opt_4}}   & 0.372 & 4.027 & 5.370     & \multicolumn{1}{c|}{1.160}           & 0.577 & 8.356 & 5.430     & 2.017           \\
				\multicolumn{1}{c|}{}                     & \multicolumn{1}{c|}{PUNet~\cite{punet}} & 0.525 & 4.601 & 5.900     & \multicolumn{1}{c|}{1.123}           & 0.544 & 6.072 & 6.841     & 0.773           \\
				\multicolumn{1}{c|}{}                     & \multicolumn{1}{c|}{PUGAN~\cite{pugan}} & 0.224 & 3.973 & 1.907     & \multicolumn{1}{c|}{1.014}           & 0.243 & 4.394 & 2.341     & 0.892           \\
				\multicolumn{1}{c|}{}                     & \multicolumn{1}{c|}{MPU~\cite{mpu}}   & 0.251 & 1.624 & 1.669     & \multicolumn{1}{c|}{1.086}           & 0.528 & 6.243 & 2.112     & 1.685           \\
				\multicolumn{1}{c|}{}                     &
				\multicolumn{1}{c|}{PU-GCN~\cite{pu-gcn}}   & 0.220 & 1.611 & 1.581     & \multicolumn{1}{c|}{0.966}           & 0.333 & 4.303 & 2.115     & 0.829           \\
				\multicolumn{1}{c|}{}                     &
				\multicolumn{1}{c|}{Ours}  & {\color{red}0.120} & {\color{red}1.606} & {\color{red}1.562}     & \multicolumn{1}{c|}{{\color{red}0.807}}           & {\color{red}0.238} & {\color{red}3.511} & {\color{red}2.009}     & {\color{red}0.760}           \\ \hline
				\multicolumn{1}{c|}{\multirow{5}{*}{8x}}  & \multicolumn{1}{c|}{EAR~\cite{opt_4}}   & -     & -     & -         & \multicolumn{1}{c|}{-}               & -     & -     & -         & -               \\
				\multicolumn{1}{c|}{}                     & \multicolumn{1}{c|}{PUNet~\cite{punet}} & 2.527 & 7.078 & 4.261     & \multicolumn{1}{c|}{0.958}           & 2.438 & 6.971 & 4.441     & 1.039           \\
				\multicolumn{1}{c|}{}                     & \multicolumn{1}{c|}{PUGAN~\cite{pugan}} & -     & -     & -         & \multicolumn{1}{c|}{-}               & -     & -     & -         & -               \\
				\multicolumn{1}{c|}{}                     & \multicolumn{1}{c|}{MPU~\cite{mpu}}   & 0.160 & 3.821 & 1.665     & \multicolumn{1}{c|}{0.732}           & 1.676 & 6.084 & 2.727     & 0.859           \\
				\multicolumn{1}{c|}{}                     &
				\multicolumn{1}{c|}{PU-GCN~\cite{pu-gcn}}   & 0.162 & 3.833 & 1.650     & \multicolumn{1}{c|}{0.741}           & 1.700 & 6.079 & 2.790     & 0.891           \\
				\multicolumn{1}{c|}{}                     &
				\multicolumn{1}{c|}{Ours}  & {\color{red}0.081} & {\color{red}2.338} & {\color{red}1.414}     & \multicolumn{1}{c|}{{\color{red}0.539}}           & {\color{red}0.142} & {\color{red}3.727} & {\color{red}2.447}     & {\color{red}0.528}           \\ \hline
				\multicolumn{1}{c|}{\multirow{5}{*}{16x}} & \multicolumn{1}{c|}{EAR~\cite{opt_4}}   & -     & -     & -         & \multicolumn{1}{c|}{-}               & -     & -     & -         & -               \\
				\multicolumn{1}{c|}{}                     & \multicolumn{1}{c|}{PUNet~\cite{punet}} & 0.460 & 4.561 & 5.891     & \multicolumn{1}{c|}{0.658}           & 0.480 & 6.143 & 4.112     & 0.804           \\
				\multicolumn{1}{c|}{}                     & \multicolumn{1}{c|}{PUGAN~\cite{pugan}} & 0.093 & 4.552 & 2.578     & \multicolumn{1}{c|}{0.491}           & 0.107 & 4.594 & 3.030     & 0.430           \\
				\multicolumn{1}{c|}{}                     & \multicolumn{1}{c|}{MPU~\cite{mpu}}   & 0.138 & 1.914 & 1.804     & \multicolumn{1}{c|}{0.489}           & 0.294 & 6.182 & 2.472     & 0.519           \\
				\multicolumn{1}{c|}{}                     &
				\multicolumn{1}{c|}{PU-GCN~\cite{pu-gcn}}   & 0.106 & 1.845 & 1.776     & \multicolumn{1}{c|}{0.479}           & 0.288 & 5.887 & 2.456     & 0.495           \\
				\multicolumn{1}{c|}{}                     &
				\multicolumn{1}{c|}{Ours}  & {\color{red}0.047} & {\color{red}1.692} & {\color{red}1.578}     & \multicolumn{1}{c|}{{\color{red}0.440}}           & {\color{red}0.097} & {\color{red}3.519} & {\color{red}2.170}     & {\color{red}0.421}           \\ \hline
			\end{tabular}%
            }
	\end{center}

\end{table}

%% file: tab/ablation.tex
\begin{table}[t]
	\caption{Quantitative comparison of the network using with or without proposed key components. \textit{Baseline} is the vanilla simple network. \textit{Full pipeline} is the proposed network.}
	\label{tab:ablation}
	\vskip -0.05in
	\begin{center}
		\begin{small}
				\begin{tabular}{c|cccc}
					\hline
					Metric($10^{-3}$)      & CD    & HD    & P2F   & Uniformity \\ \hline
					\textit{Feature BP}         & 0.256 & 6.113 & 3.234 & 0.667           \\
					\textit{Coordinate BP}      & 0.137 & 4.122 & 3.001 & 0.486           \\
					\textit{Position Embeddings} & 0.238 & 5.879 & 3.112 & 0.640           \\ \hline
					\textit{Baseline}           & 0.322 & 6.547 & 3.556 & 0.714           \\ \hline
					\textit{Full pipeline}      & 0.097 & 3.519 & 2.170 & 0.427           \\ \hline
				\end{tabular}
		\end{small}
	\end{center}
	\vskip -0.3in
\end{table}